\documentclass[sigconf]{acmart}

\AtBeginDocument{%
  \providecommand\BibTeX{{%
    \normalfont B\kern-0.5em{\scshape i\kern-0.25em b}\kern-0.8em\TeX}}}

\usepackage{float}
\usepackage{subcaption}
\copyrightyear{2023}
\acmYear{2023}
\setcopyright{rightsretained}

\acmConference[AIES '23]{AAAI/ACM Conference on AI, Ethics, and Society}{August 8--10, 2023}{Montréal, QC, Canada}
\acmBooktitle{AAAI/ACM Conference on AI, Ethics, and Society (AIES '23), August 8--10, 2023, Montréal, QC, Canada}
\acmDOI{10.1145/3600211.3604669}
\acmISBN{979-8-4007-0231-0/23/08}

\makeatletter
\patchcmd{\maketitle}{\@copyrightpermission}{
   \begin{minipage}{0.3\columnwidth}
     \href{http://creativecommons.org/licenses/by/4.0/}{\includegraphics[width=0.90\textwidth]{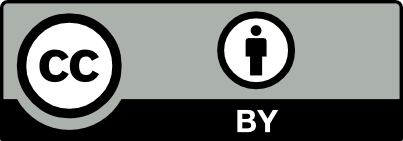}}
   \end{minipage}\hfill
   \begin{minipage}{0.7\columnwidth}
     \href{http://creativecommons.org/licenses/by/4.0/}{This work is licensed under a Creative Commons Attribution International 4.0 License.}
   \end{minipage}
  
   \vspace{5pt}
}{}{}

\begin{document}

\title{User Tampering in Reinforcement Learning Recommender Systems}

\author{Atoosa Kasirzadeh}
\affiliation{%
  \institution{The Alan Turing Institute \\ The University of Edinburgh}
  \city{Edinburgh}
  \country{United Kingdom}
}
\email{atoosa.kasirzadeh@ed.ac.uk}

\author{Charles Evans}
\affiliation{%
  \institution{The Australian National University}
  \city{Canberra}
  \country{Australia}
}
\email{charlie.evans@warwick.ac.uk}


\begin{abstract}
In this paper, we introduce new formal methods and provide empirical evidence to highlight a unique safety concern prevalent in reinforcement learning (RL)-based recommendation algorithms -- 'user tampering.' User tampering is a situation where an RL-based recommender system may manipulate a media user's opinions through its suggestions as part of a policy to maximize long-term user engagement. We use formal techniques from causal modeling to critically analyze prevailing solutions proposed in the literature for implementing scalable RL-based recommendation systems, and we observe that these methods do not adequately prevent user tampering. Moreover, we evaluate existing mitigation strategies for reward tampering issues, and show that these methods are insufficient in addressing the distinct phenomenon of user tampering within the context of recommendations. We further reinforce our findings with a simulation study of an RL-based recommendation system focused on the dissemination of political content. Our study shows that a Q-learning algorithm consistently learns to exploit its opportunities to polarize simulated users with its early recommendations in order to have more consistent success with subsequent recommendations that align with this induced polarization. Our findings emphasize the necessity for developing safer RL-based recommendation systems and suggest that achieving such safety would require a fundamental shift in the design away from the approaches we have seen in the recent literature.
\end{abstract}



\maketitle

\section{Introduction}


Recommender systems, also known as recommendation systems, are algorithms designed to sift through vast collections of data to identify and suggest entities that are particularly relevant to a specific user or group \citep{bobadilla13}. These systems have been extensively deployed in a variety of domains, including entertainment, retail, and social media, where they curate suggestions for movies, music, and merchandise, as exemplified by platforms such as Netflix, YouTube, and Twitter. A particularly significant application of these systems is in news and social media platforms where they curate relevant content for users. In the context of this paper, we focus on these as \emph{media recommender systems}.

A popular approach to deploying recommender systems is to treat the recommendation problem as a Markov Decision Process (MDP) and applying reinforcement learning (RL) to the recommendation task. Although the potential of this approach was recognized theoretically two decades ago  \citep{shani05, taghipour07, taghipour08}, the more recent emergence of `Deep RL' -- notably, its ability to handle larger, more complex recommendation problems -- has reignited applied interest \citep{liu18, zhao18b, zhao18a}. In response, researchers have begun exploring the applicability of Deep RL-based recommendations within the news and social media sector \citep{zheng18, shahbazi20}. This body of work has shown a significant increase in user engagement compared to the deployment of the recommendation problem using two other prominent methods: (i) `static' machine learning approaches \citep{liu10, garcin12, bobadilla13, lu15, anandhan18} and (ii) contextual Multi-Armed Bandit approaches \citep{li10, tang14, tang15, zeng16}.

Advancements in RL techniques have enabled the large-scale implementation of RL-based recommender systems. Major social media platforms, such as Facebook, have already begun integrating these systems into their frameworks \citep{gauci18, liu20}. This integration raises crucial safety and ethical question: How can we identify potential harms arising from the use of RL-based recommender systems, and what measures can we take to mitigate them?

The social implications of media recommender systems have received significant attention in recent years \citep{khwaja19, abdollahpouri20, milano20, patro20, stray20}. A comprehensive review of the topic identifies six key areas of concern: biased/unfair recommendations, encroachment on individual autonomy and identity, opacity, questionable content, privacy, and social manipulability and polarization \citep{milano20}. This paper focuses on the last of these concerns: social manipulability and polarization. In particular, we elucidate the potential harms and risks of social manipulation and polarization posed by RL-based media recommendation systems. Given the growing ubiquity of RL-based social media in our daily lives, we argue that these concerns necessitate immediate scrutiny. We begin our discussion with a review of the primary literature on this subject.

\citet{russell19, russell19youtube} suggests that a specific issue of social manipulation and polarization can arise when the recommendation problem is viewed as an MDP, and an RL approach is used to resolve it. The primary concern here is that an RL-based recommender system might learn to make recommendations based not solely on the user's current interests and beliefs, but also on the long-term influence these recommendations could have on the user. This approach could result in altering the user's interests and beliefs over time. Various recent studies have investigated this issue from different angles

\citet{krueger20} connect this problem to a concrete probabilistic concept known as auto-induced distributional shift (ADS). ADS pertains to the capacity of an RL agent to learn independently how to shift the opinion distribution among users to its own objective advantage. Such a shift might involve (i) enticing a larger proportion of users who are simpler to provide recommendations for, or (ii) modifying the preferences and behaviors of the current user base. \citet{krueger20} explore methods to mitigate such unwanted manipulation in example problems including media recommendation systems. However, the learning algorithms employed in their studies are not reflective of the current state-of-the-art algorithms with which this study is concerned, as they consider the emergence of ADS in the context of population-based training with multilayer perceptrons, rather than deep RL. Our paper attempts to address this gap.

\citet{carroll21} examine the dangers of RL-based recommenders insofar as the inducement of opinion shifts -- or ADS -- is concerned. The authors formulate mitigation strategies against these dangers by revising the optimization objective of the RL agent. \citet{farquhar22} also examine the media recommendation problem and explore a different set of mitigation strategies, relying upon a novel extension of the classic MDP model and Causal Inference theory to limit the RL agent's ability to identify opportunities for manipulating users. Our paper differs from these two works in that, while the previous authors focus on principles for how safer RL-based recommenders \emph{could} be developed, we delve deeper into the question of \emph{why they must} be urgently developed with urgency. We substantiate our point through both mathematical and computational demonstrations.

In this paper, we make two core contributions to the literature on safe and ethical media recommender systems. Our first contribution is the formalization of the concept of `user tampering' as a potential safety issue that could arise in RL-based media recommenders. User tampering reflects the concern that a recommender system may learn that manipulation of a user's preferences, opinions, and beliefs via the recommendation of certain content has beneficial outcomes for its ability to maximize its reward function in the longer term.\footnote{In the rest of this paper, we use the terms `preferences', `opinions', `beliefs, and `interests' interchangeably.} 

User tampering is formalized using the Causal Influence Diagram techniques, proposed by \citet{everitt21b}, to extract the specific mechanisms enabling RL-based recommenders to learn such manipulating strategies. Unlike previous research, our formalization directly engages with the state-of-the-art algorithmic designs featured in recent RL-based recommendation literature. Our second contribution is an experimental demonstration of user tampering in recommender systems. In particular, we design a simulation study capturing a simple media recommendation problem. We show that a standard Q-learning algorithm can learn to exploit user tampering by developing a policy for making recommendations that affect our simulated users' content preferences. While our simulation occurs on a significantly smaller scale than a real recommendation problem scenario, its novelty is relevant because it aims to replicate a known cause of opinion shift in social media users. Thus, our simulation study computationally affirms that user tampering is a crucial ethical and safety concern which must be taken seriously when designing and deploying RL-based media recommender systems.

The rest of this paper is structured as follows. In Section 2, we formulate the media recommendation problem as a Markov Decision Process. We then introduce a causal model of the recommendation problem, which we think can be representative of a large subset of current leading recommender systems. Section 3 introduces user tampering formally. We draw on Casual Influence Diagram techniques to identify problematic behavioural incentives in our proposed problem formulation. We then use these techniques to articulate why mitigation strategies applicable to similar tampering problems cannot apply successfully to user tampering. Section 4 introduces our simulation study and its results. Finally, Section 5 concludes the paper.

\section{Modelling the Media Recommendation Problem}
\label{sec:modelling}

In this section, we achieve two goals. First, we present a formulation of the media recommendation problem as a Markov Decision Process (MDP).\footnote{For an introduction to Markov Decision Processes, see \citet{puterman1990markov}.} Second, we employ Causal Influence Diagrams (CIDs) to identify relevant causal relationships among specific variables within this model. Our formulation aims to maintain the MDP as general as possible, while integrating design insights from recent developments in the implementation of RL-based media recommender systems \citep{zheng18, shahbazi20}.

\subsection{The MDP formulation of the media recommendation problem}
\label{sec:formula_mdp}

We begin by constructing an MDP model that represents the media recommendation problem, one that aligns with those commonly employed in recent RL-based recommendation literature. This approach is informed by a recent survey, which outlines the cutting-edge of research into RL-based recommendation algorithms, as detailed by \citet{afsar22}. Our proposed MDP formulation $\langle S, A, R, T, \gamma \rangle$ includes a series of well-founded assertions, specifically relevant to a reasonable MDP formulation of a media recommendation problem. These assertions are as follows:

\begin{itemize}
\item $S$ denotes a set of states. A state $s \in S$ can represent a variety of structures, but primarily, it encodes information about the performance of recent recommendations from the recommender. As noted in \citet{afsar22}, this form of state representation is broadly applicable to the various methods of modeling media recommendation problems in recent literature. Dominant approaches since the mid-2000s tend to represent the state based on recent positive user-content interactions \citep{taghipour07, taghipour08}. This approach to representation is also observed in the Deep RL literature that has gained attention over the past five years \cite{afsar22}. As a concrete example among many, consider the state represented by a collection of $|n \times m|$ data points, which capture users' aggregate clicks on recommended items across $n$ categories and through $m$ different time frames of recent history (e.g. the last 1 hour, 6 hours, 1 day, etc.).\footnote{This example is very similar to the approach taken by \citet{zheng18} to represent states.} The inclusion of user-behavioral information is crucial in state representation; without it, the theoretical advantages of using RL could be compromised. In order to develop policies that not only capture current opportunities for reward but also anticipate future ones, it is necessary to incorporate user-behavioral information in the state representation.
    
\item $A$ denotes a set of actions. As \citet{afsar22} observe, recent studies show a substantial consistency in how actions are modeled, with an action signifying either a single item or a collection of items recommended to a user. Practically speaking, actions could manifest as an $n$-dimensional vector, representing the properties of a piece of content (e.g., an article) across $n$ dimensions for user recommendation. This concept can be broadened to interpret an action as the recommendation of a fixed-size bundle of content (e.g., a set of articles) to a user, since the individual content units can be integrated into the vector.

\item $R$ denotes a reward function. $R$ maps an agent's activity to numerical values indicative of the `goodness' of these activities. Typically, recommender systems use observable engagement metrics, such as a clicks or 'likes', as a basis for these rewards. The form of the function $R$ may vary based on the specific implementation and the definition of actions and the state space. It can be represented in several ways, including $R : S \rightarrow \mathbb{R}$ or $R : S \times A \rightarrow \mathbb{R} $. In the case of recommender systems, as engagement is included in the state representation and will be updated at each step, a function of the form $R : S \rightarrow \mathbb{R}$ is generally sufficient.

\item $T$ denotes a transition function. $T$ calculates the probability of an agent arriving at a specific `successor state' $s'$ after taking a specific action $a$ from its current state $s$. Typically, a transition function is formulated as $T: S \times A \times S \rightarrow [0,1]$.

\item $\gamma \in \mathbb{R}$ denotes a discount factor for future rewards. $\gamma$ encapsulates the balance between the value assigned to immediate rewards and those expected in the future.
\end{itemize}

Given these five types of assertion, a media recommendation problem can be modeled as follows: An agent takes an action $a_t$ ($a_t \in A$) at time $t$. This action transitions the system from the current state $s_t$ ($s_t \in S$) to a subsequent state $s_{t+1}$ ($s_{t+1} \in S$), with the probability $T(s_t,a_t,s_{t+1})$. Following this transition, the agent receives a reward, denoted as $R(s_{t})$. Subsequently, another action is chosen at time $t+1$, and the process continues. During this sequence of actions and rewards, the influence of the discount factor $\gamma$ is consistently factored in.

\subsection{Extracting a CID from the MDP}
\label{sec:extracting_cid}

The Causal Influence Diagram (CID) is a modeling technique central to our formalization of user tampering \citep{heckerman95, everitt21b}. This technique has recently seen increased application in analyzing the potential incentives driving RL agents' behaviour \citep{armstrong20, everitt21b, everitt21a}. We start by briefly describing the basic building blocks of CIDs and then will provide an illustration of CIDs within the context of the media recommendation problem (Figure 1).

CIDs are structured as directed acyclic graphs. CIDs are specified by three node types, representing different variables within the problem at hand. These nodes are (i) Decision Nodes, (ii) Structural Nodes, and (iii) Utility Nodes. As depicted in Figure 1, Decision Nodes are shown as squares, Structural Nodes as circles, and Utility Nodes as diamonds. The Decision Nodes stand for variables that receive an assigned value at the point of decision.

Both Structural Nodes and Utility Nodes symbolize probability distributions over the possible values a variable might take. However, they do this in different ways: Structural Nodes represent distributions over possible state variable values, whereas Utility Nodes represent distributions over possible rewards. A directed edge from a node $X$ to a node $Y$ can be interpreted as follows. If $Y$ is a Utility or Structural Node, then the value of the random variable $Y$ is conditional on the value of $X$. In such instances, a solid line depicts the edge. If $Y$ is a Decision Node, then the value of $X$ represents the information available to the agent at the decision-time of $Y$. The edge, in this case, is illustrated with a dashed line.\footnote{The reader is encouraged to refer to Everitt et al. \citep{everitt21b} for a more comprehensive understanding of the CID modeling technique.}

In the context of RL-based media recommendation, using a CID-based analysis for agent incentives offers several advantages when compared to alternative methods like purely statistical analyses. Firstly, CIDs, and the notion of an instrumental goal in these graphs (as introduced in Section~\ref{sec:ut}), are uniquely equipped to handle \emph{causation}, not just \emph{mere correlation} between variables. This is pivotal in media recommendation, where the causal dependency relations are crucial: We can show recommenders' ability to cause increased user engagement via their actions' causal effects on users' preferences and opinions. Secondly, CIDs allow us to abstract from extraneous information about the specifics of RL algorithm implementations and instead focus on the core causal mechanisms shared among them. This abstraction is particularly beneficial in providing a space for formally discussing causal properties of various potential implementations simultaneously, as opposed to a statistically analysis of the results achieved by each unique implementation.

Note that throughout this paper, for the given CIDs, our figures only depict a subgraph of three time steps from the entire diagram. This simplification is intended to capture the main structure of CIDs, without overly complicating the visualization. Consequently, these results can be readily generalized to scenarios involving more than three time steps.

\begin{figure}
    \centering
    \includegraphics[width=0.35\textwidth]{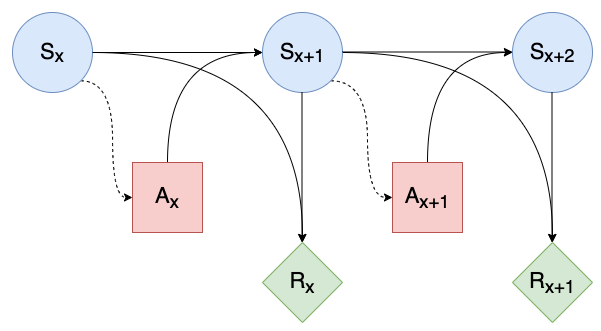}
    \caption{A naive CID of the media recommendation problem.}
    \label{fig1:naive_CID}
\end{figure}

Let us begin with a simplified model of a media recommendation problem. If we were to naively model an MDP's causal structure as a CID, without additional considerations, we would end up with a representation akin to that shown in Figure~\ref{fig1:naive_CID}. At a specific time step $x$, the distribution over possible current states is represented by $S_x$. The actual value of the state at time $x$ is the only piece of information available to the agent for its action selection at $A_x$. The distribution of possible states in $S_{x+1}$ is subsequently determined by $T$, given $S_x$ and $A_x$. Finally, $R_x$ denotes the distribution over the reward value achieved from the action taken at $A_x$. We have assumed an interpretation of the reward function as $R : S \times S \rightarrow \mathbb{R}$, wherein the reward is determined by comparing two successive states. This arrangement offers adequate information to infer the success of the most recent action or recommendation.

A simple thought experiment demonstrates that this naive CID underspecifies the causal relationships in the actual problem due to neglecting key variables external to the MDP. Consider the following scenario. Alice and Bob are two university students who recently created accounts on a media platform. Thus far, both have been recommended the same three articles about student politics at their university and both have clicked on the three articles. Within our problem specification, it is quite plausible that the states of the system have been identical thus far from the recommender agent's perspective. Yet, suppose Bob clicks on the articles because his friends feature in the cover photos of the three articles, whereas Alice's clicks stem from a genuine interest in politics, including student politics. If the next recommendation for both Alice and Bob -- denoted as $A_{x+1}$ -- is an article on federal politics, the distribution over possible states at $S_{x+1}$ is the same. In this state, Alice is more likely to be observed engaging with this content.

This thought experiment illustrates the need for an exogenous random variable to the MDP to model any external causal effects potentially introducing media recommendations. We think this variable can capture the relevance of a specific user's hidden impactful opinions to whom the agent recommends media content. We represent this exogenous variable by $\theta^T$ for considering it in our causal modeling framework.

That is to say, the exogenous variable that is the user's interests not captured by her observed behavior (such a click or like) at time $x$, is represented as $\theta^T_x$.\footnote{We do not enforce any Markov assumptions on $\theta^T_x$: it may depend on the values of the variable at multiple, or even all, previous time steps.} The key point is the potential causal relationship between $\theta^T_x$ and $S_{x+1}$. 

As the simple example above demonstrates, an appropriate explanation for the distribution over states $S_{x+1}$ cannot be achieved without considering the possible effect of $\theta^T_x$.

This potential link cannot be easily removed by any practical redesign of the state space. Moreover, it is crucial to recognize that an influence link also exists between $A_x$ and $\theta^T_{x+1}$. This reflects the intuitive idea that a user's information consumption may modify their interests over time. Although $\theta^T$ is exogenous, we are not proposing a precise model that explains \emph{how} $A_x$ affects the distribution over possible values of $\theta^T_{x+1}$. Rather, we are acknowledging the potential for a causal dependency via this influence link.

\begin{figure}
    \centering
    \includegraphics[width=0.5\textwidth]{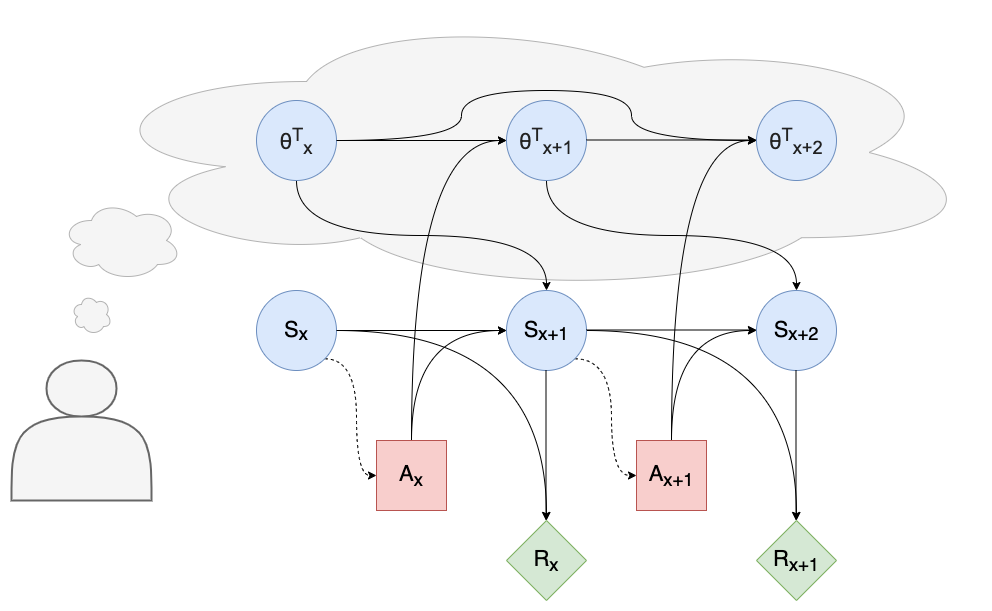}
    \caption{A CID of the media recommendation problem, extended to include the exogenous variable affecting state transitions.}
    \label{fig2:tf_only}
\end{figure}

By incorporating the exogenous variable $\theta^T$ in our model of media recommendation, we can revise Figure~\ref{fig1:naive_CID} to the CID depicted in Figure~\ref{fig2:tf_only}. We believe that this better captures the actual causal dynamics at play in the media recommendation MDP. We would like to note that similar causal structures for the recommendation process have been suggested in previous work \citep{jiang19}. Nonetheless, these were not framed within the CID context. Our approach, in contrast, not only integrates these structures into the CID framework, but also facilitates in-depth graphical analysis of the media recommendation systems, which we will elaborate on in the following section.

The exact design of the MDP can lead to variations in the CID formulated here; for one such variants see Appendix~\ref{sec:mdp_variation}.\footnote{In this Appendix, we provide an example of the implied causal structure when the designer opts to broaden the reward function to incorporate observations not captured in the state representation.} However, these variations do no impact the role or influence of $\theta^T$ and its links from the preceding action to the succeeding state remains part of the model's causal structure. As our forthcoming analysis will specifically focus on these causal relationships, the CID depicted in Figure~\ref{fig2:tf_only} offers a sufficiently general representation for our needs going forward.

\section{User Tampering}
\label{sec:ut}

In this section, we use the CID outlined in the previous section (Figure~\ref{fig2:tf_only}) to examine a primary safety concern related to RL-based media recommendation systems: the potential for user manipulation and polarization. In particular, we introduce and formalize the phenomenon of user tampering. This refers to the possibility of an RL-based recommender system strategically manipulating a user's opinions via its suggestions, aiming to maximize long-term user engagement.

After introducing the concepts of `instrumental goals' and `instrumental control incentives,' we demonstrate within the CID framework that an instrumental goal exists for the agent to manipulate the expected value of the exogenous variable, $\theta^T$. This provides a concrete and formal interpretation of the user tampering safety concern.

\subsection{Instrumental Goals and Control Incentives}

According to \citet{everitt21b}, an `instrumental goal' is conceptualized as an outcome that serves as a means to achieve the ultimate goal of obtaining a reward. Speaking in causal terms, an agent possesses an instrumental goal to cause an event if: (1) the agent is able to cause the event and (2) the event, in turn, results in an increase in the agent's expected observed reward. One of the key benefits of using CIDs in the modeling of media recommendation systems is that CIDs provide us with conceptual tools for the examination of instrumental goals. RL agents often harbor such instrumental goals, which assist in increasing their observed rewards.

An `instrumental control incentive' (ICI) is a property displayed in the graphical models of CIDs, as introduced by \citet{everitt21b}. An ICI is said to be present on a Structural Node $X$ if it is located along a path in the CID that originates at a Decision Node and concludes at a Utility Node. This essentially implies that the choice of action at the Decision Node can alter the expected utility at the Utility Node \emph{through} affecting the distribution over values at $X$.

ICIs bear significant implications for user tampering due to their capacity to graphically indicate either the possible presence or the categorical absence of an \emph{instrumental goal} in certain events within a RL problem \citep{everitt21a}. An RL agent is said to possess an instrumental goal to influence the distribution at a Structural Node $X$ in a certain way if it has an ICI on $X$ and that particular influence increases the expected reward accumulated by the agent. Essentially, the agent must have both the ability and a motive to influence the distribution at $X$.

\subsection{Formalizing User Tampering}

\begin{figure}
    \centering
    \includegraphics[width=0.45\textwidth]{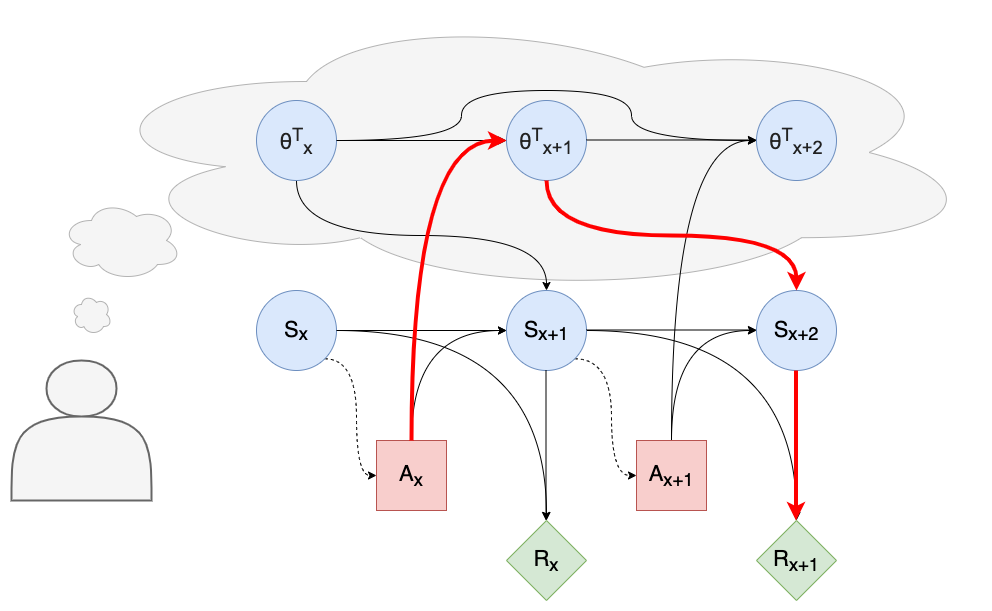}
    \caption{An annotated version of the media recommendation CID for state-based rewards. An example of an undesirable causal path introducing an instrumental control incentive on $\theta^T_{x+1}$ is shown in bolded red.}
    \label{fig5:tf_only_ann}
\end{figure}

In the CID presented in Figure~\ref{fig5:tf_only_ann}, there is a subset of Structural Nodes upon which instrumental goals are \emph{desirable}. These nodes represent the set of random state variables, denoted as $\{S_t | t \in \mathbb{N}\}$. The term `desirable' here means that we, as the problem's designers or framers, `want' the RL agent to shift the probability distribution at these nodes towards `good' states which maximize reward. In our problem, these are states where many of recent recommendations have been favorably received by the user. As such, any path in the CID from a Decision Node to a Utility Node passing exclusively through random state variables (e.g. [$A_x \rightarrow S_{x+1} \rightarrow R_{x}$], or [$A_x \rightarrow S_{x+1} \rightarrow S_{x+2} \rightarrow R_{x+1}$]) only involves intended and safe instrumental goals.

However, in the CID, there exist additional paths from Decision to Utility Nodes. Specifically, there are paths which involve the exogenous random variables -- for example, [$A_x \rightarrow \theta^T_{x+1} \rightarrow S_{x+2} \rightarrow R_{x+1}$]. This pathway is illustrated in Figure~\ref{fig5:tf_only_ann}. An ICI is clearly present on $\theta^T_{x+1}$ or on any other variable in $\{\theta^T_t | t \in \mathbb{N}\}$ appearing in similar pathways.\footnote{Page size constraints prevent us from displaying larger CID subgraphs, but note that longer paths can also be identified containing similar instrumental goals. For example, paths of the form [$A_x \rightarrow \theta^T_{x+1} \rightarrow \theta^T_{x+2} \rightarrow ... \rightarrow \theta^T_{x+n} \rightarrow S_{x+n+1} \rightarrow R_{x+n}$], or [$A_x \rightarrow \theta^T_{x+1} \rightarrow S_{x+2} \rightarrow S_{x+3} \rightarrow ... \rightarrow S_{x+n} \rightarrow R_{x+n-1}$] are feasible.}

Given these conditions, if an agent can secure higher rewards by tailoring recommendations to a user with particular interests (represented by $\theta^T$), then the agent may have an instrumental goal to influence $\theta^T$ accordingly, potentially leading to greater long-term expected rewards. Essentially, the presence of an ICI on at least one node in $\{\theta^T_t\ | t \in \mathbb{N}\}$ in the CID establishes the graphical prerequisite for user manipulation to emerge as an instrumental goal for an RL agent. If such an instrumental goal is attainable -- meaning the agent can boost its expected reward by influencing users' interests -- then we can expect that an advanced RL agent would likely learn to exploit this instrumental goal, rendering user tampering a `learnable' phenomenon. We can thus define user tampering as follows.

\bigbreak
\emph{\textbf{Definition 1}. User tampering is a `learnable' phenomenon for an RL-based media recommendation algorithm iff it has an instrumental goal to affect at least one of the variables in $\{\theta^T_t\ | t \in \mathbb{N}\}$}.
\bigbreak

Importantly, however, an instrumental goal affecting some variable in $\{\theta^T_t\ | t \in \mathbb{N}\}$ does not inherently mean that a given RL agent will necessarily learn to affect the user in a way that increases its expected reward. It simply means that the agent \emph{has the potential} to learn this behavior. So, the learnability of user tampering in a certain model is a necessary, but not a sufficient, condition for user tampering to actually occur in an RL agent's learned policy. To clarify this distinction, it is beneficial to introduce a second definition of user tampering that separates our discussion of its theoretical learnability from the examination of its practical manifestations in a specific recommender's policy.

\bigbreak
\emph{\textbf{Definition 2}. An RL-based media recommendation algorithm `exploits' user tampering iff there exists a state $s_t$ such that $\pi(s_t) = a_t$ and $\pi'(s_t) \neq a_t$, for the algorithm's actual learned policy $\pi$, and the hypothetical policy $\pi'$ that the same learning process would have produced in a world where each action has no causal link to the user's subsequent preferences, i.e., $A_t \perp \!\!\! \perp \theta^T_{t+1} $.}
\bigbreak

Informally, this is to say that the learned policy makes a different recommendation in some possible state of the problem than what it would make in a hypothetical scenario where recommendations had no causal impact upon the user's preferences.

In the rest of this section, we contrast our proposal to a different form of `tampering' in RL, known as `Reward Function (RF)-tampering.' Despite some apparent similarities, the RF-tampering is quite distinct from a causal perspective. This distinction rules out the transfer of promising solutions in the literature -- particularly those proposed by Everitt et al. \citep{everitt21a} -- from the context of RF-tampering to that of user tampering.

\subsection{User Tampering's Differences from RF-tampering}

Reward function (RF)-tampering refers to a specific safety issue where an RL agent has one or several undesirable instrumental goal(s) to affect variables \emph{within} its own reward function. This aims to alter the way in which certain states are evaluated by the function \citep{armstrong20, everitt21a}. Detailed analysis on this issue is scant, but it is suggested that the high-level concerns of social manipulation and polarization might be classified under the category of RF-tampering \citep{everitt21a}. At a certain level, this assertion seems intuitive -- the user and their behavior often mirror a `reward function' for the recommendation system, since the user's response ultimately decides whether a recommendation is rewarded. Hence, one might expect that tampering with the user would constitute a kind of `reward' tampering. However, our earlier discussion reveals that this assumption is inaccurate.

In the media recommendation problem, the reward function is an explicitly defined function that maps concrete outcomes within the state space, such as clicks or likes, to numerical rewards. Here, the user essentially forms part of the problem environment, with their behavior contributing to the environment's dynamics. Our earlier definition of 'user tampering' is not a form of 'reward tampering.' Instead, it more accurately represents 'transition tampering.' This perspective is acknowledged by \citet{everitt21a}, but not dealt with nearly as extensively as their work on reward tampering.

To situate our argument, we begin with a brief synopsis of the characteristics of any problem in which RF-tampering (and its associated CID) may occur. In such a problem:

\begin{itemize}
\item The reward function can be expressed as $R(S; \theta) : S \times \mathbb{R}^N \rightarrow \mathbb{R}$, where $\theta$ represents some `parameters' of the reward function other than states or actions. These parameters are distinct from the states or actions in the model, and should not be confused with $\theta^T$ from our model; we use $\theta$ here to be consistent with the notation of \citet{everitt21a}.

\item A specific `intended' value of $\theta$ exists, represented as $\theta_*$. This value remains static unless a change is introduced to it at some point by an external process. In other words, its value is independent of any actions undertaken by the RL agent.
\item The agent models $\theta_*$ and updates that model based on its experiences. At each time step $t$, the agent's distribution over possible values of $\theta_*$ is represented as $\theta_t$.
\item The agent is able to influence the distribution $\theta_{t+1}$ with its action $a_t$.
\item The rewards observed by the agent at time step $t$ are defined as $R(S_t; \theta_t)$, rather than $R(S_t; \theta_*)$.
\end{itemize}
    
The crux of the RF-tampering problem is that the agent has an instrumental goal to alter its own model of the reward parameters such that it rewards certain states more positively than what $\theta_*$ would actually generate.

Note, however, that the MDP and associated CID representation of the media recommendation do not conform to this description on several points. Particularly, in the media recommendation problem:

\begin{itemize}
\item $\theta^{T}$ is not a hidden parameter to the \emph{reward} function, but instead to the \emph{transition} function.
\item $\theta^{T}$ is \emph{not} independent of the agent's actions. While it fulfils a similar conceptual role as $\theta_*$ in RF-tampering in that it represents an `intended' parameter, it is nonetheless subject to the effects of the agent's recommendations. This is why there is a causal link from $A_t$ to $\theta^{T}_{t+1}$.

\item In the problem of media recommendation, it is generally not attempted to estimate the distribution $\theta^{T}_t$ at a given time step $t$. Rather, the state space contains an implicit estimation of the intended parameters in the form of recorded user behavior. This contributes to the causal link between $\theta^T_t$ and $S_{t+1}$. This does not imply that there \emph{cannot} be an attempt to model this aspect explicitly. For example, \citet{carroll21} attempt this explicit modeling in their work aimed at mitigating user manipulation. However, as the current industry standards and R\&D trends outlined by \citet{afsar22}, it is standard not to do so.
\end{itemize}

\begin{figure}
    \centering
    \includegraphics[width=0.3\textwidth]{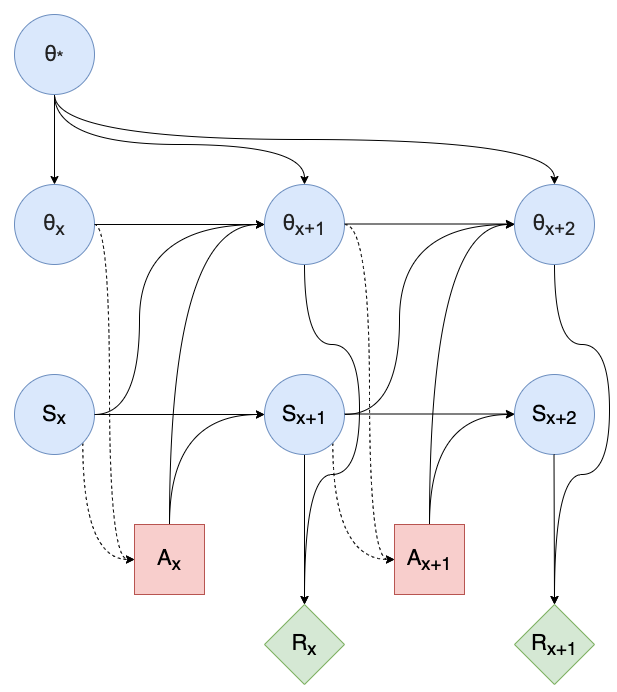}
    \caption{CID Representation of an RF-Tampering-susceptible problem.}
    \label{fig4:everitt_cid}
\end{figure}

It may help the reader to consult and compare the diagram in Figure~\ref{fig4:everitt_cid}, where we have recreated the CID given in \citet{everitt21a} to represent an RF-tampering-susceptible model, with the recommendation CID we constructed in Section~\ref{sec:modelling} (i.e. Figure ~\ref{fig2:tf_only}).

Several solutions to RF-tampering have been proposed by \citet{armstrong20} and \citet{everitt21a}. However, some of their underlying assumptions cannot be transferred to our case of transition tampering. Although one proposed solution could be adapted theoretically, its implementation demands a resolution to as-of-yet unresolved questions in the literature, as we elaborate on below.

The solutions that cannot transfer to transition tampering are what \citet{everitt21a} refer to as Time-inconsistency-considering, Direct learning, and Counterfactual agents.\footnote{Readers may refer to their work for further details about each of these concepts.} At their core, these solutions aim to eliminate the agent's instrumental goals to tamper with its own model of the reward function through hidden parameters. 

For our purposes, it is important to understand that these solutions are predicated on the assumption of `uninformativity.' This means that no causal links can exist between the ground truth hidden parameters at one time step, and the distribution over the state space at the next time step (meaning graphically, in our model, no directed arrows $\theta^T_t \rightarrow S_{t+1}$). As we have discussed, this is an unattainable quality in the media recommendation problem, because it undermines the entire purpose of approaching recommendation with RL. To reiterate, if the state representation contains any measure of previous recommendation successes, 'uninformativity' is not achieved. Conversely, if it does not, an RL algorithm fails to learn as it becomes incapable of drawing associations between its recommendations and their respective impacts on overall reward. Consequently, we must disregard this set of RF-tampering solutions as a feasible transfer of ideas to the user tampering problem.

Another solution proposed is the concept of a time-inconsistency-ignoring agent, which does not necessitate uninformativity, thus providing potentially more promise. The basic premise of this solution -- as it would apply to user tampering -- is to initially model the user's content preferences explicitly, and subsequently rewarding the agent based on the engagement we would \emph{expect} its recommendations to receive according to this model, rather than the user's actual engagement. This idea was proposed by \citet{everitt21b}, who believed that this approach eliminates the ICI, and the instrumental goal, to manipulate user preferences.

However, this solution reveals complex, unresolved problems: How can we learn a sufficiently detailed model of a user's content preferences? How can we simulate organic evolution in these preferences without making the model dependent on the user's actual behavior? For example, if a user of their own volition develops an interest in politics, resulting in increased engagement with political content when occasionally recommended, it would be a poor service if the system continued to seldom recommend such content due to its reliance on an outdated preference model.

A deeper understanding of how and why preferences shift `naturally' apart from content consumption, and how to distinguish this from changes induced by tampering, is essential. Although recent work by \citet{carroll21} makes progress in this area, further research is required.

The takeaway for us, in this paper, is that there is no `quick fix' to tackle user tampering issues, extrapolated from other research on RL tampering problems. Coupling this with our prior discussions about the surge in popularity of RL recommenders and the dominance of a problem framework that significantly enables learnable user tampering, a disconcerting image of the current safety of RL media recommendation starts to surface. To further illustrate this, we will present computational results in the next section, showing exploitation of user tampering in simulated scenarios.

\section{Computational experiments}
\label{sec:experiment}

In this section, we empirically analyze the user tampering phenomenon formalized in the previous section. First, we introduce a simple media recommendation problem, which involves simulated users and a user tampering incentive, derived from recent empirical findings concerning polarization on social media. Second, we present a Q-learning agent designed to mimic the Deep Q-learning algorithms employed in recent media recommendation research, training it within this environment \citep{zheng18, shahbazi20}. Our findings show that the policy it learns significantly exploits user tampering to maximize rewards.

\subsection{Problem Formulation and Environment Setup}

Consider the following scenario where a recommender agent sequentially offers $h$ `political post/article' recommendations to a user. At each time step $t$ ($0 \leq t \leq h$), the agent can select one of three available `sources' for recommendation. The first source present consistently left-wing in its perspective, the second offers consistently a centrist viewpoint, and the third consistently showcases a right-wing stance.

Furthermore, we assume the definition of the exogenous parameter $\theta^T$ introduced in Section~\ref{sec:modelling}. Recall that the agent does not explicitly model this variable ($\theta^T$ is an exogenous variable). We define $\theta^T_t$ as a tuple of three probabilities as of time $t$, i.e. $\Theta^T = \{(\theta^{T^L}, \theta^{T^C}, \theta^{T^R}) \in \mathbb{R}^3 \: | \: \forall x \in \{L,R,C\}.\theta^{T^x} \in [0,1]\}$. For some arbitrary user, their probability $\theta^{T^L}$ represents their probability of clicking an article from the left-wing source \emph{if it is recommended}; the same can be said of $\theta^{T^C}$ for the centrist source, and $\theta^{T^R}$ for the right-wing source. We say that a user is initially (t=0) `right-wing' iff $\theta^{T^R}_0 > \theta^{T^C}_0 \land \theta^{T^R}_0 > \theta^{T^L}_0$ , and `left-wing' iff $\theta^{T^L}_0 > \theta^{T^C}_0 \land \theta^{T^L}_0 > \theta^{T^R}_0$. Finally, we include a simple environmental dynamic whereby users who are recommended content from a source that is politically opposed to their viewpoint gradually become more polarized in favor of their own political bias. This concept is underpinned by recent studies exploring user polarization on social media. These studies provide evidence that exposure to a high volume of content from the politically opposite side can often amplify user polarization \citep{bail18, bail21}.

We would like to emphasize that our model of polarization is greatly simplified and is not intended to model the intricate details of the polarization phenomenon described in applied social media in the previously cited works. Indeed, our primary goal is not to simulate the effects of polarization in painstaking detail, but rather to construct an environment which allows the hypothesized effect of user tampering to be tested given the potential for polarization. In order to accomplish this, we consider potential causal effect an agent could use as part of their instrumental goal. Even though our model is simplified, its dynamics remain rooted in authentic sociological findings.

The detailed definition $\langle S,A,T,R,\gamma \rangle$ of the media recommendation MDP, as well as the precise implementation of the `polarization' effect we have just described, is provided in Appendix~\ref{sec:app_c}. Next, we train a Q-learning agent in this environment and show that it learns to perform user tampering on our simulated users.

\subsection{Recommender Simulation}
\label{sec:recsim}

To computationally operationalize the model described previously, we need to establish some additional specifications. We assign a value of 30 to $h$, while the probabilities that define the exogenous variable $\theta^T$ are restricted to a maximum value of 0.75.\footnote{The authors imposed this arbitrary limitation to prevent users from becoming so 'polarized' that they would engage with every post from a source that mirrors their viewpoints. This seemed an extreme, and thus unrealistic, outcome that could undermine the plausibility of our simulation.} We then introduce $p$, defined as the 'polarization factor.' The polarization factor represents a user's subsequent likelihood of engaging with content from their aligned source, after having a post from an opposing source recommended to them.

In the context of our experiment, a population of five `users' with varying preference profiles. This include:

\begin{itemize}
    \item A `strong left' user with $\theta^T_0$ = (0.4, 0.1, 0.1)
    \item A `moderate left' user with $\theta^T_0$ = (0.3, 0.25, 0.1)
    \item A `centrist' user with $\theta^T_0$ = (0.2, 0.4, 0.2)
    \item A `moderate right' user with $\theta^T_0$ = (0.1, 0.25, 0.3)
    \item A `strong right' user with $\theta^T_0$ = (0.1, 0.1, 0.4)
\end{itemize}

We train a Q-learning agent in this environment. Each episode starts with the selection of a user at random from the population to provide the initial $\theta^T$ value.\footnote{Our implementation, including a pre-trained recommender agent, is available on GitHub: https://github.com/chevans-lab/user-tampering.} Non-deep Q-learning was used for training, in spite of deep Q-learning being the more viable approach at industrial scales; this was a deliberate choice, because unlike deep Q-learning, non-deep Q-learning provably converges towards the optimal policy for the problem.\footnote{Since we wanted to test whether the agent was able to find a \emph{better} policy by exploiting user tampering than it could otherwise achieve, Deep Q-learning was an inappropriate choice for the experiment as there was no way to guarantee that it would not converge on a good, safe policy even when a better, user tampering policy was available.} Nonetheless, to maintain alignment with practical Deep RL application, we modeled the state space in a parameterized manner suitable for such algorithms.


For each of the five users in our population, we provide two plots based on 10000 evaluation episodes with the user, using the policy learned from the aforementioned training process. 

Respectively, these two plots estimate the following. (i) The probability of the learned policy selecting each action at every problem time-step. This is determined by taking the per-episode average frequency of each choice. (ii) The expected reward accumulated up to and including each time step $t$, $ 0 \leq t \leq h$. To provide context, we plot this scenario against the expected reward accumulated by two different kinds of recommenders. 

(a) The first recommender makes uniformly random recommendations at each time step. (b) The second recommender follows a simple multi-armed bandit-esque policy, which provides a `baseline' of a good policy. This policy makes random recommendations for the first third of the episode, but then operates like a multi-armed bandit, always recommending from the source that has the highest mean reward in the episode thus far.

\begin{figure}
    \centering
    \begin{subfigure}[b]{0.5\textwidth}
        \centering
        \includegraphics[width=\textwidth]{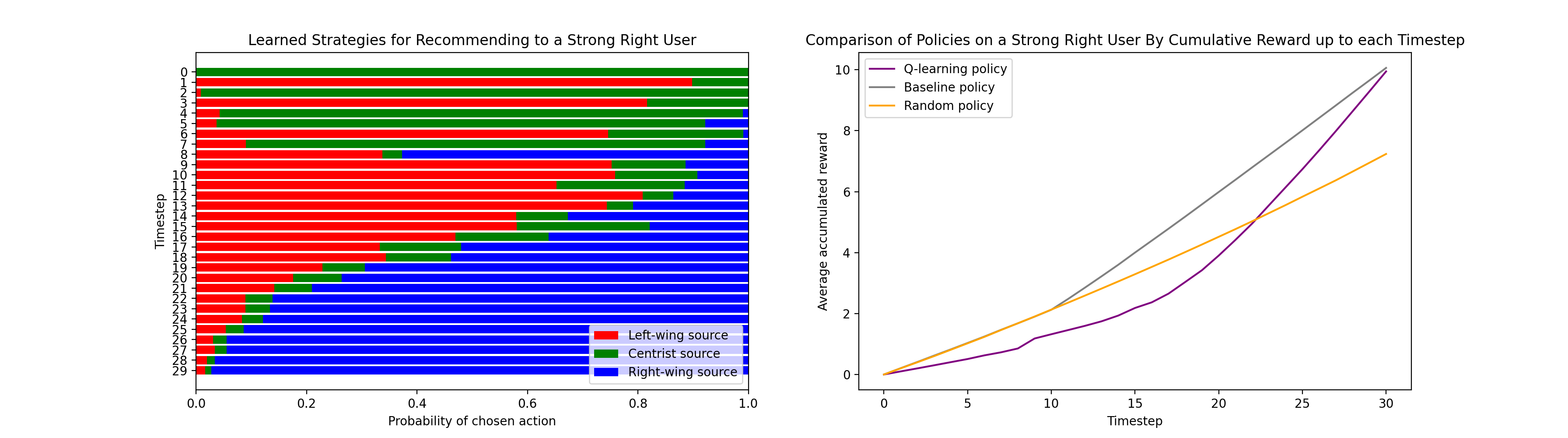}
        \caption[Network2]%
        {{\tiny The `Strong Right' user.}}    
        \label{fig4A:SR}
    \end{subfigure}
    \vskip\baselineskip
    \begin{subfigure}[b]{0.5\textwidth}  
        \centering 
        \includegraphics[width=\textwidth]{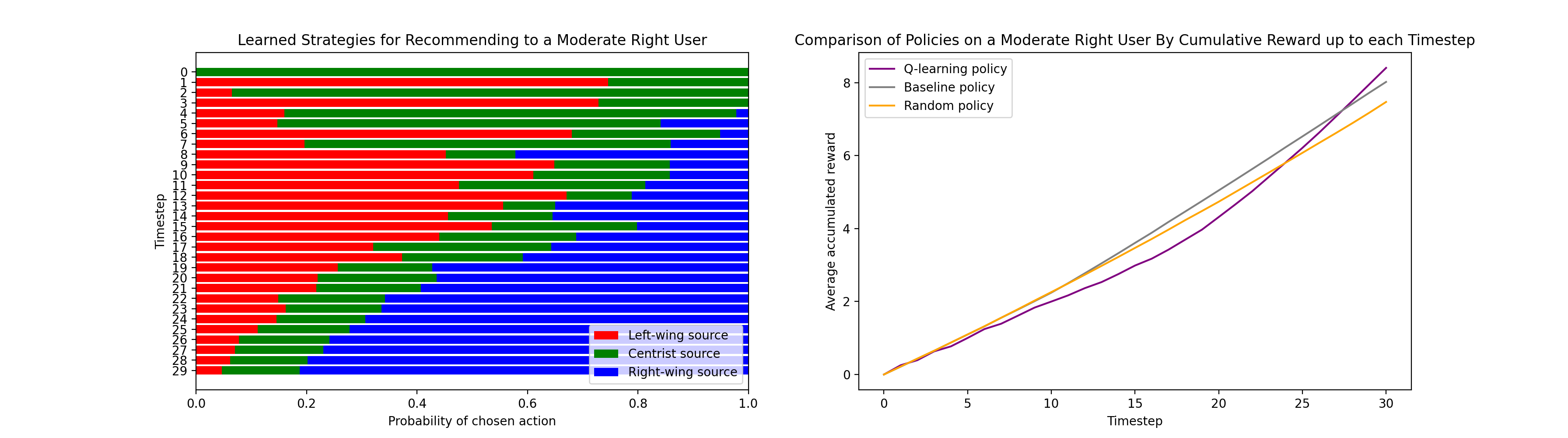}
        \caption[]%
        {{\tiny The `Moderate Right' user.}}    
        \label{fig4B:MR}
    \end{subfigure}
    \vskip\baselineskip
    \begin{subfigure}[b]{0.5\textwidth}   
        \centering 
        \includegraphics[width=\textwidth]{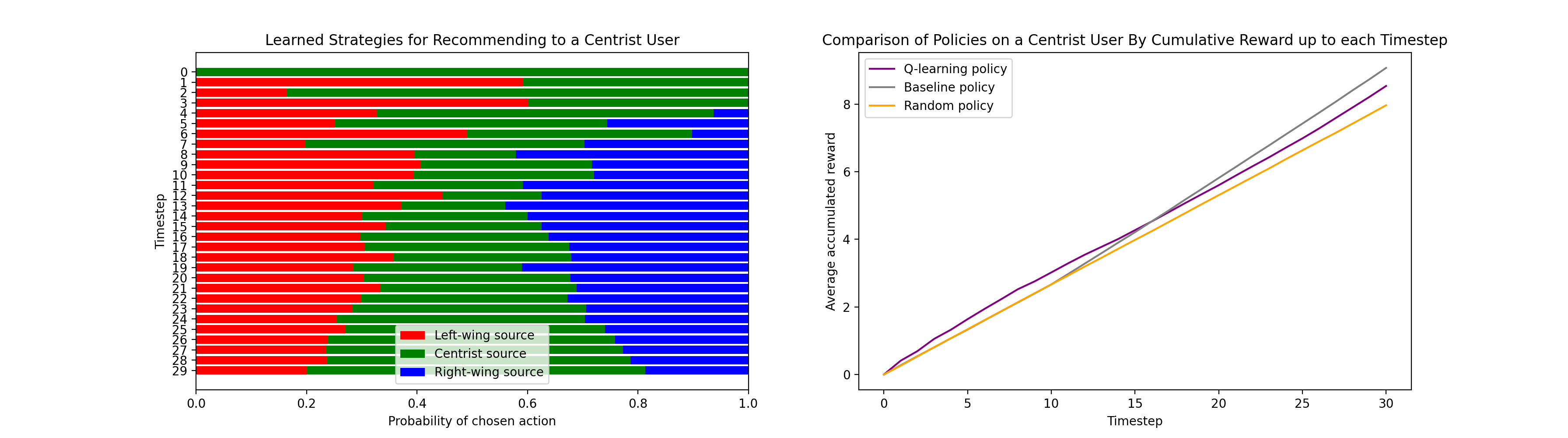}
        \caption[]%
        {{\tiny The `Centrist' user.}}    
        \label{fig4C:C}
    \end{subfigure}
    \vskip\baselineskip
    \begin{subfigure}[b]{0.5\textwidth}   
        \centering 
        \includegraphics[width=\textwidth]{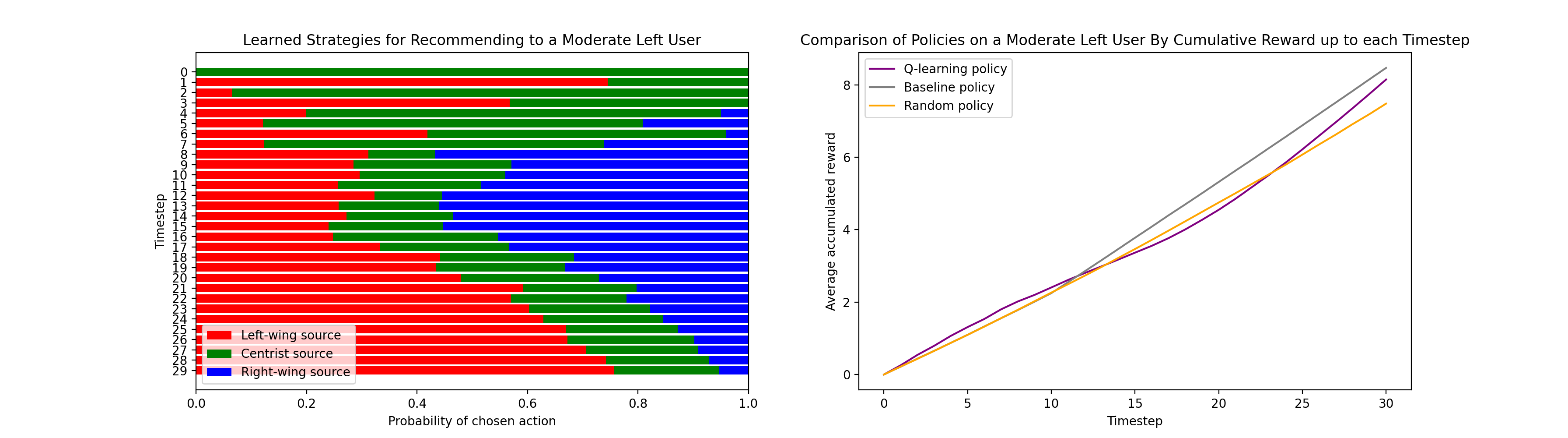}
        \caption[]%
        {{\tiny The `Moderate Left' user.}}    
        \label{fig4D:ML}
    \end{subfigure}
    \vskip\baselineskip
    \begin{subfigure}[b]{0.5\textwidth}   
        \centering 
        \includegraphics[width=\textwidth]{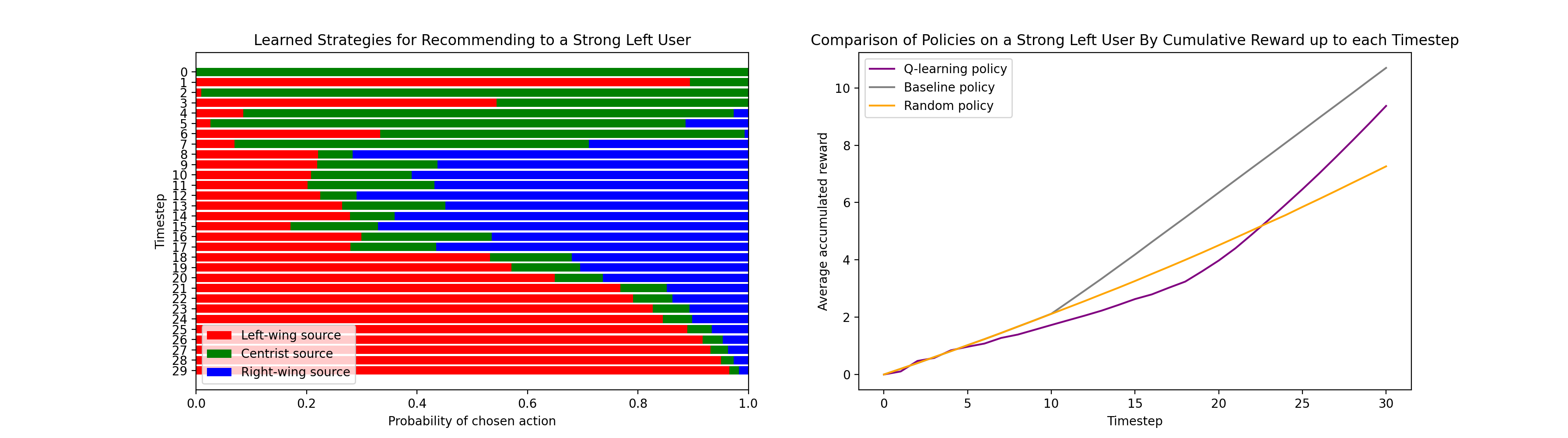}
        \caption[]%
        {{\tiny The `Strong Left' user.}}    
        \label{fig4E:SL}
    \end{subfigure}
    \caption{Evaluation of the policy learned with Q-learning for each member of our sample user population.}
    \label{fig4AToE:plots}
\end{figure}

Figure~\ref{fig4AToE:plots} illustrates the plots for each of the simulated users as specified earlier. These results exhibit multiple interesting properties. 

First, the exploitation of user tampering by the learned policy is apparent for all users, with the exception of the Centrist. Directing attention to the strategy plots of the two 'left-wing' users reveals a particularly dominant strategy that evolves as follows:

\begin{itemize}
\item The recommender attempts to profile the user and their preferences. This is achieved by assessing their response to centrist and left-wing content -- predominantly during the first quarter of the episode.
\item The recommender primarily recommends right-wing content \emph{in spite of} its low expected reward. This will tamper with the user's preferences and increase the expected reward from subsequent left-wing recommendations -- mainly during the second quarter of the episode.
\item The recommender predominantly recommends left-wing content to the (now more) left-wing leaning user. This maximizes the high expected rewards that action will now offer -- mainly during the second half of the episode.
\end{itemize}

Despite the low expected rewards associated with right-wing content and the learnability of user tampering in this context, the recommender system is observed to strongly favor right-wing content initially, only to later shift to left-wing recommendations for the rest of the episode. This apparent exploitation of user tampering is noteworthy. Moreover, the inverse behavior is learned for right-wing users -- it suggests that the model is not merely attempting to polarize all users towards the left, but is instead crafting a nuanced policy to identify and exploit the causal link between its actions and the user's exogenous variable. This inference is further supported by the policy observed for 'centrist' users.

Further evidence to this effect is given by the policy for the `centrist' user -- here, the data clearly indicates that the recommender recognizes its actions hold no discernible causal influence over these users, preventing any feasible user tampering. Consequently, the system's recommendations align closely with the initial preferences of these users.

Second, the agent heavily exploits user tampering even though we were able to generate similar cumulative rewards with our crude `baseline' policy. This adds weight to the safety concerns with respect to user tampering. It indicates that there exist other policies which do not exploit user tampering (although they may make a handful of `polarizing' recommendations by chance) and which offer similar rewards to the one that the recommender learned; nonetheless, over several iterations of retraining, the policy consistently converged to the policy we have presented here (with small natural variations). This implies that in this environment, the unsafe policy is not only learned occasionally, but presents a likely direction of convergence for the learning algorithm.

It is also worth establishing that the exploitation of user tampering in the learned policy was robust to simulated users not encountered during training. We generated the same policy plots for the recommender over 10000 evaluation episodes spent recommending to each user in a new, `unseen' population: an `extremely left' user with $\theta^T_0$ = (0.5, 0.05, 0.05), an `extremely right' user with $\theta^T_0$ = (0.05, 0.05, 0.5), a `left anti-centrist' user with $\theta^T_0$ = (0.35, 0.05, 0.2), and a `right anti-centrist' user with $\theta^T_0$ = (0.2, 0.05, 0.35).

\begin{figure}
    \centering
    \begin{subfigure}[b]{0.225\textwidth}
        \centering
        \includegraphics[width=\textwidth]{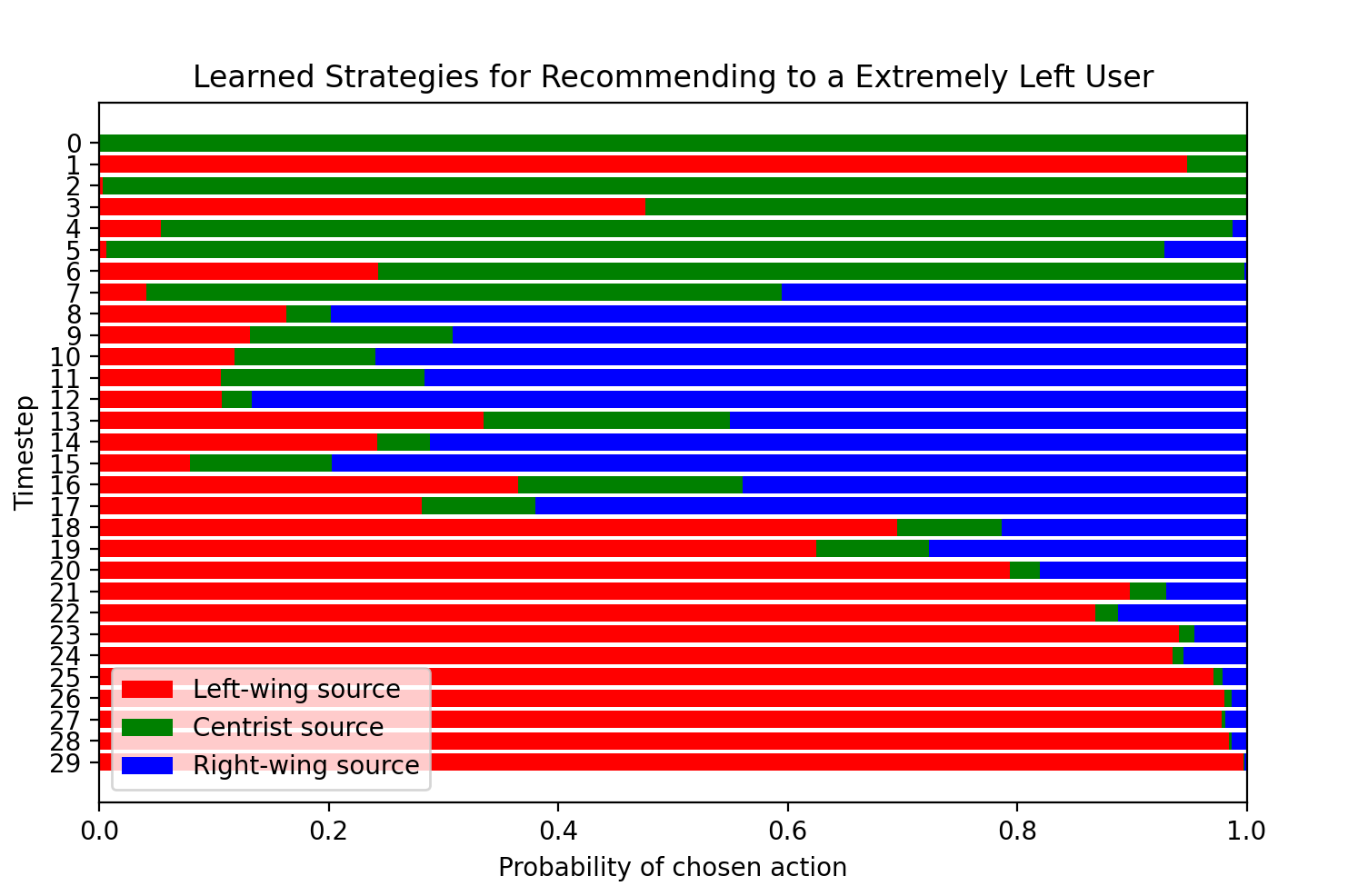}
        \caption[Network2]%
        {{\tiny The `Extremely Left' user.}}    
        \label{fig5A:EL}
    \end{subfigure}
    \begin{subfigure}[b]{0.225\textwidth}  
        \centering 
        \includegraphics[width=\textwidth]{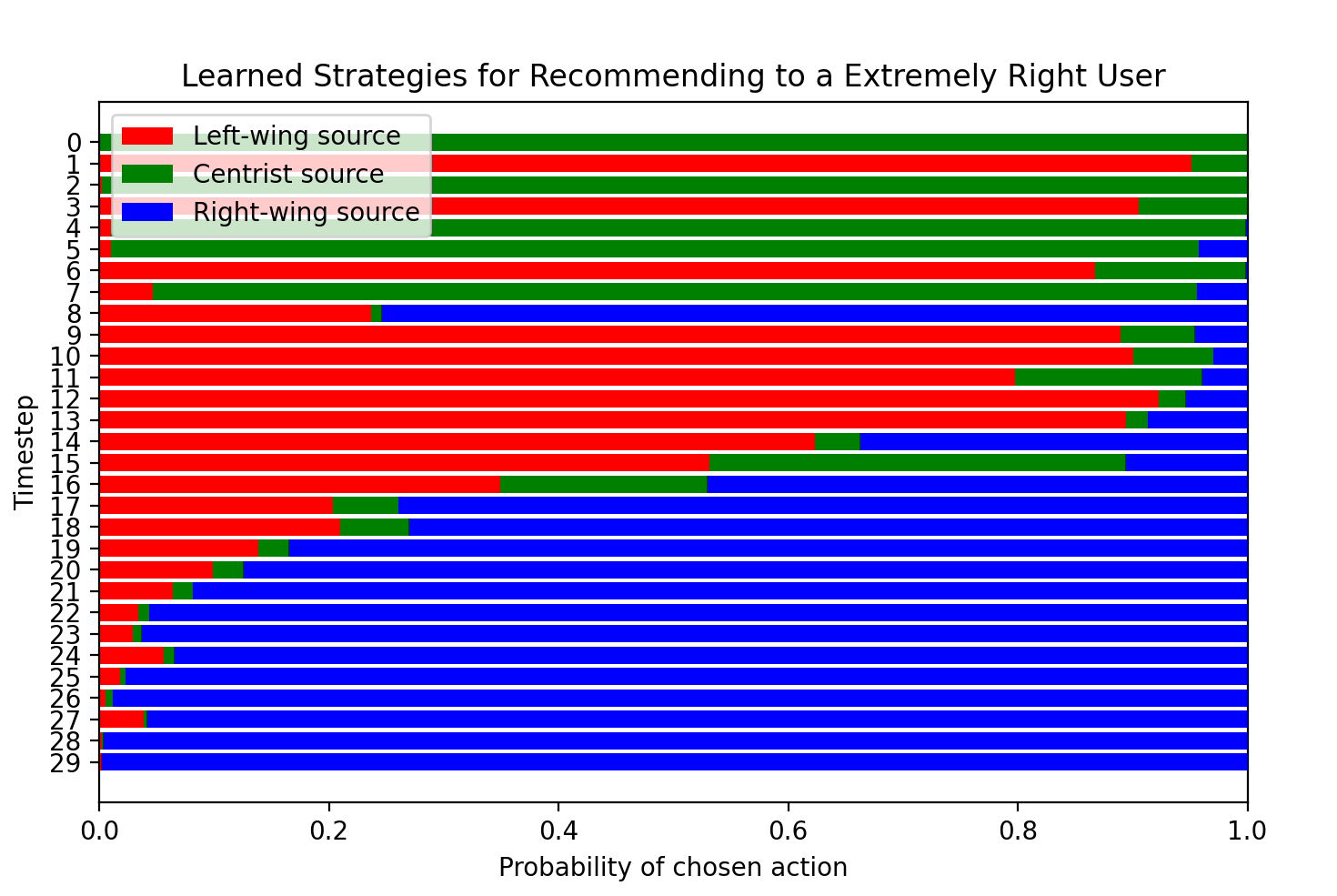}
        \caption[]%
        {{\tiny The `Extremely Right' user.}}    
        \label{fig5B:ER}
    \end{subfigure}
    \vskip\baselineskip
    \begin{subfigure}[b]{0.225\textwidth}   
        \centering 
        \includegraphics[width=\textwidth]{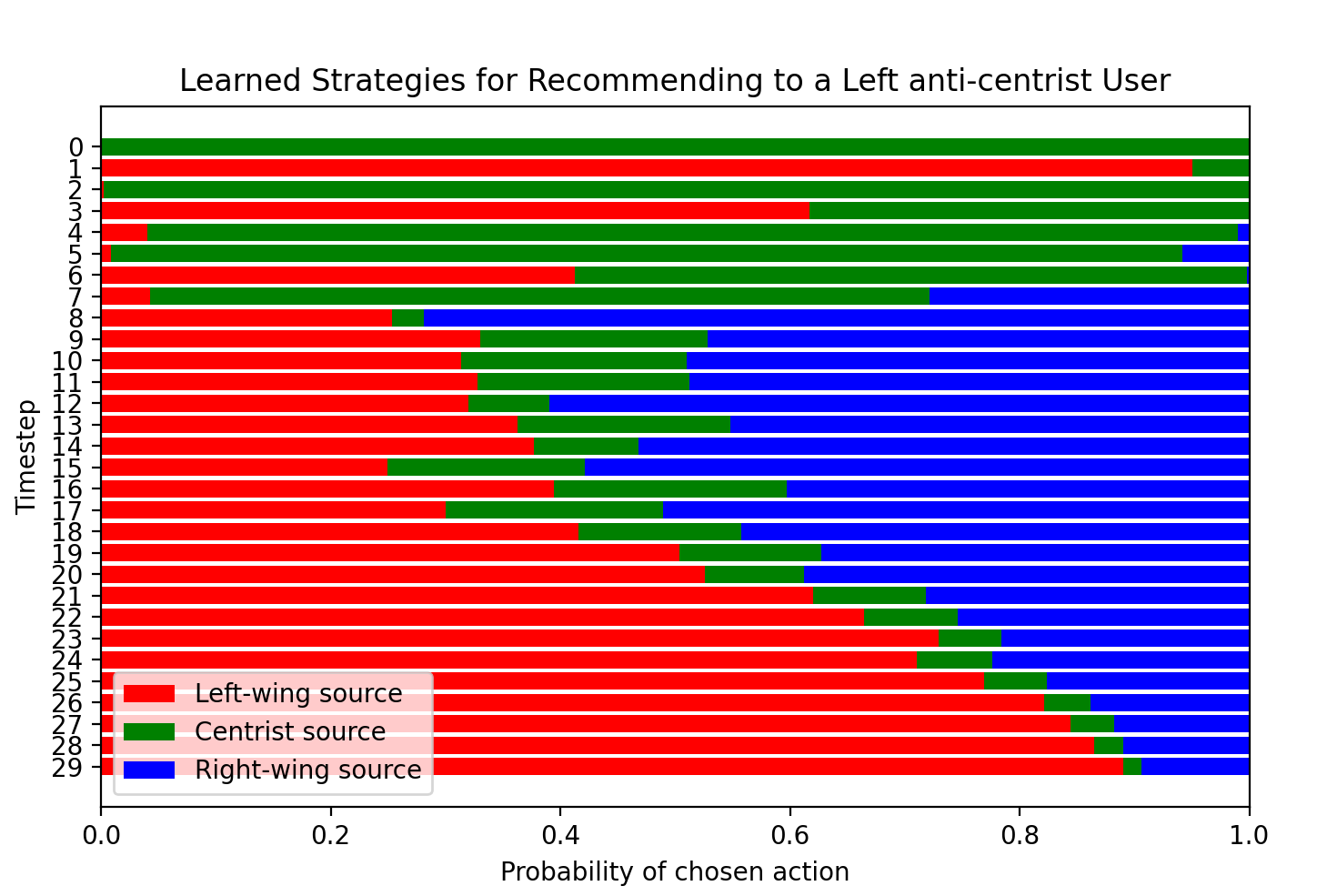}
        \caption[]%
        {{\tiny The `Left anti-centrist' user.}}    
        \label{fig5C:LAC}
    \end{subfigure}
    \begin{subfigure}[b]{0.225\textwidth}   
        \centering 
        \includegraphics[width=\textwidth]{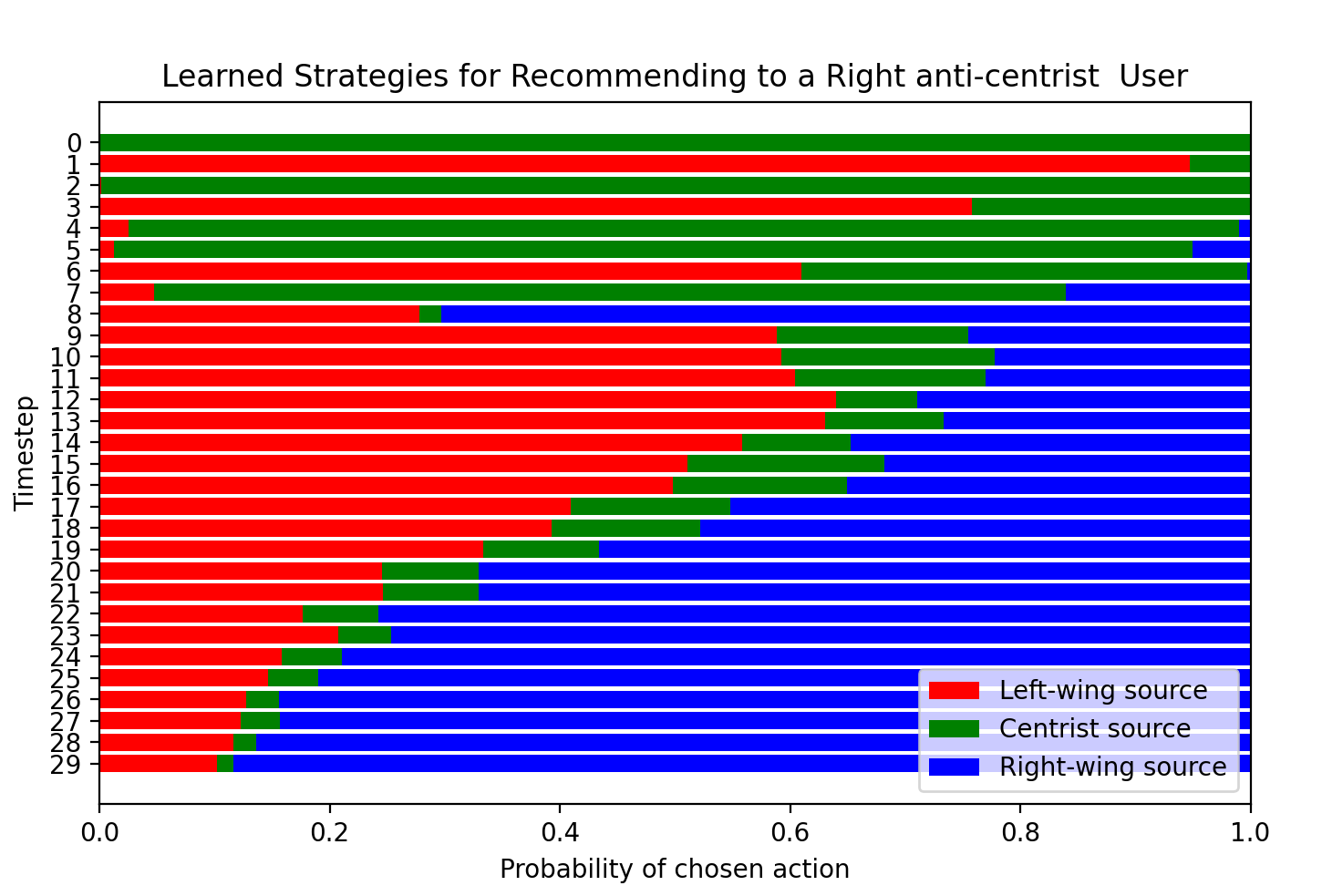}
        \caption[]%
        {{\tiny The `Right anti-centrist' user.}}    
        \label{fig5D:RAC}
    \end{subfigure}
    \caption{Action probabilities at each time-step for each user in the `unseen' population.}
    \label{fig5AToD:unseen_user_plots}
\end{figure}

These results are shown in Figure~\ref{fig5AToD:unseen_user_plots}. Although these specific users were never encountered during training, the same unsafe strategies appear here; the three phases of user profiling, then polarization, and finally preference satisfaction are clearly visible.

These results support the previous section's claims that user tampering is learnable for commercially dominant methods of designing RL media recommender systems, and strengthen the implications of this by showing that it is, at least according to our small-scale simulation, very much exploitable. Taken in combination with the lack of immediately available remedies, this should raise significant safety concerns about the use of the current state of the art in RL recommendation on public media platforms.

\section{Discussion and Conclusion}
\label{sec:conc}

This paper has substantiated concerns about the risks of emergent RL-based recommender systems with respect to user manipulation and polarization. We have formalized these concerns as a causal property -- user tampering. We have demonstrated the possibility of isolating and identifying user tampering within a formalization of a recommendation system's implicit causal model. We have discussed why the learnability of user tampering is practically uniformly present amongst leading RL recommender systems, and why research into similar RL tampering problems cannot easily be adapted to redesign RL recommendation systems to be safer. Moreover, we have demonstrated computational results for a simple simulation environment which we designed inspired by recent research on social manipulation and polarization. We have shown that a Q-Learning-based recommendation algorithm can consistently learn a policy of exploiting user tampering -- which, in our discussion, took the form of the algorithm explicitly polarizing our simulated users. We argued that our demonstration of user tampering phenomenon points to the potential unethical and troubling problems in real-world media recommendation systems. Due to a combination of technical and pragmatic limitations on what could be done differently in RL-based recommender design, we urge significant caution in the deployment of RL media recommendation systems until commercially and computationally viable adaptations of these algorithms that explicitly protect against the possibility of user tampering have emerged.

To this end, the findings in this paper motivate further work in two distinct areas; increasing understanding of the possibilities of the user tampering phenomenon in practice, and identifying positive directions for advancement in research \& development of safer algorithms.

While this paper has formalized user tampering and demonstrated its exploitation by an RL algorithm in a simulated environment, that environment was highly abstracted relative to an industrial-scale recommendation problem. So, while it was valuable in showing that user tampering \emph{can} be exploited by an RL-based recommender system, it contributed less to our understanding of how it would manifest in a real context. While experimenting with actual users would obviously raise ethical questions, there is room for progress by simply reducing the level of abstraction present in the simulation. For example, future work can consider simulating the following scenarios:
\begin{itemize}
    \item Recommendations from a wider range of sources, which may or may not be political.
    \item Users with more complex preference profiles, and users whose preference profiles shift as a result of effects external to the recommendation environment during a recommendation episode.
    \item Causal effects of recommendations on simulated users which more authentically replicate empirically demonstrated effects on real users.
    \item Recommending content over longer episodes.
\end{itemize}

On a related note, it would also be valuable to show that results similar to ours can be replicated with a Deep Q-learning-based algorithm, given that this may more closely replicate the learning process of industrial-scale RL-based recommendation (which is predominantly done with Deep Q-learning). Extending our results with this or any of the above suggestions would further substantiate our concerns by closing the abstraction gap between these simulations and the real-world system being simulated.

With respect to identifying positive directions for recommendation research, we believe that the combination of (a) a recommender algorithm capable of learning to estimate a recommendation's effect on the success of subsequent recommendations and (b) defining success in terms of user engagement poses inevitable risks in the form of user tampering. Note, however, that this does not invalidate the premise that algorithms which have a temporally sophisticated approach of the kind described in the first point above \emph{do model the recommendation problem more effectively than static approaches}; where possible, then, this remains an attractive property to include in recommendation algorithms' implementation. We suggest that these observations could be interestingly combined with recent discussions on the notion of `multistakeholder' recommendation \citep{abdollahpouri20, patro20, stray20}. This discourse, to summarize, has pushed for a more `value-aligned' approach to recommender system design that -- without disregarding the (primary, from a business perspective) goal of user engagement -- recommendations should also reflect the interests of other stakeholders such as the creators of the content being recommended, and even society at large. For us, this raises the question; could we create a multi-stakeholder recommendation system in which the interests of the non-user stakeholders, at least, benefit from a more temporally sophisticated approach?

We suggest that approaching multi-stakeholder recommendation with an ensemble model, where each sub-model represents the interests of a (group of) stakeholder(s) and all sub-models \emph{except} for the `user-representing' sub-model are RL-based may be an exciting direction for future research. This would allow the potential benefits of the RL approach to be maximized as far as possible without introducing learnable user tampering. Without violating the pragmatic requirement that maximizing user engagement is a primary driver of the ensemble's recommendation decision, such an approach may allow us to create systems which use RL's potential in recommendation not as an enabler for user manipulation and polarization, but instead as a positive force for achieving multi-stakeholder value alignment.

\begin{acks}

We thank the {\it Humanising Machine Intelligence} grand challenge and its members at ANU for support and feedback throughout this research, as well as attendees of the 4th FAccTRec Workshop on Responsible Recommendation at RecSys 2021, members of the Causal Incentives Working Group at DeepMind, and the anonymous reviewers for their respective feedback. Special thanks are given to Lexing Xie, Tom Everitt, Sebastian Farquhar and Micah Carroll for their comments and discussion. Significant portions of this paper were written while Atoosa Kasirzadeh was at the Australian National University.
\end{acks}

\bibliographystyle{ACM-Reference-Format}
\bibliography{main}

\appendix

\section{Example Variation on the Media Recommendation MDP and CID}
\label{sec:mdp_variation}

The MDP representation of the media recommendation problem may be expanded relative to our characterization in Section~\ref{sec:modelling}, if the designer wishes to expand the reward function to account for observations that are not captured in the state representation. This would be a reasonable design choice -- for example, the state representation may only record some user behaviors such as clicks, whereas we may want to reward the agent based not only on clicks, but also on the `dwell time' of the user on the article (the time spent on the article after clicking). For generality, we demonstrate how the CID could be extended to represent this.

This firstly requires some changes and introductions to our MDP definition:

\begin{itemize}
    \item A set of observations $O$. An observation consists of some collection of metrics representing how a user observably responded to some recommendation. 
    \item An observation probability function $Z : S \times A \times O \rightarrow [0,1]$. This models the probability of making a particular observation (for example a click, but no comment) after making a certain recommendation in a certain state.
    \item An altered definition of the Reward function as $R : O \rightarrow \mathbb{R}$. This simply corresponds to the fact that the information on which rewards are predicated -- the observable user response to the content -- has now been concentrated into the one variable $o \in O$.
\end{itemize}

\begin{figure}[h]
    \centering
    \includegraphics[width=0.45\textwidth]{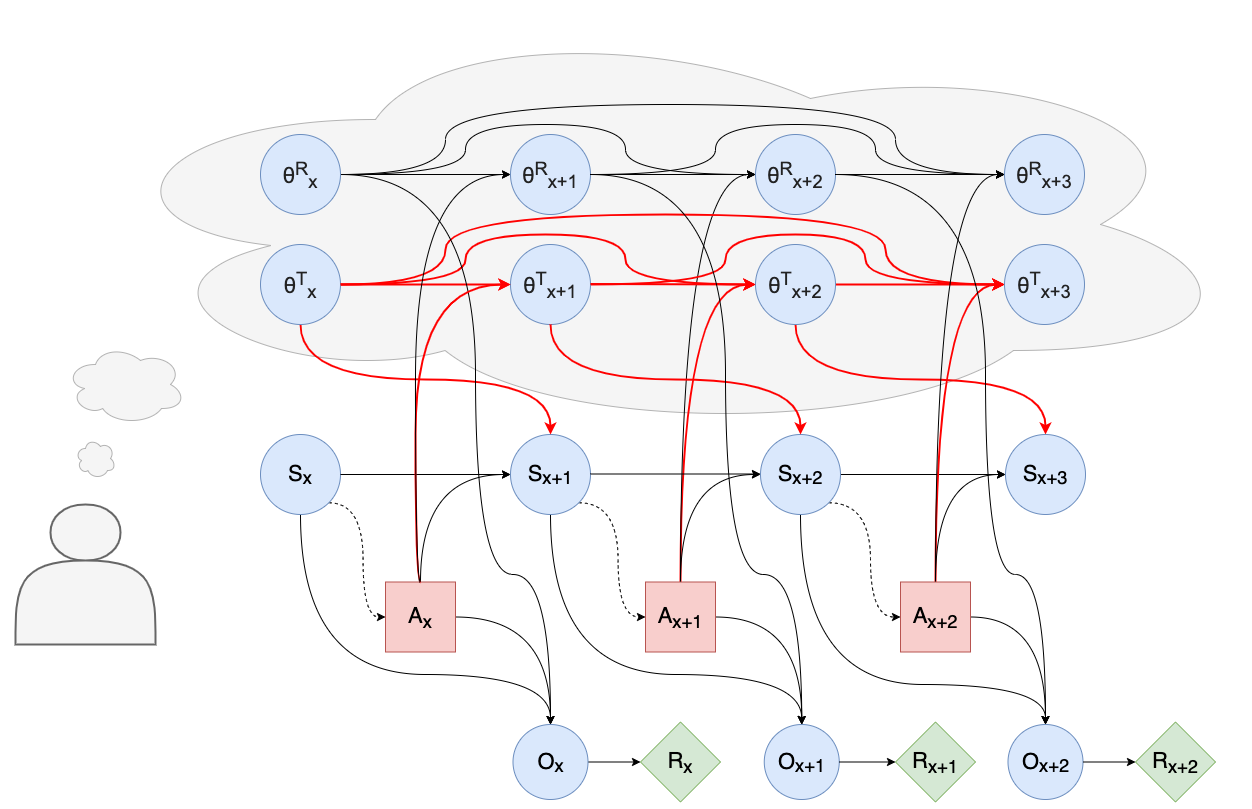}
    \caption{A CID of the media recommendation problem, extended to include an observation space and more complicated definitions of reward.}
    \label{fig3:tf_rf}
\end{figure}

We also need to make the addition of an exogenous random variable $\theta^R$ for the updated CID. This serves a highly similar purpose to $\theta^T$, except that it instead accounts for the fact that the probability of observing a certain behaviour in response to an article will intuitively change from user to user, even if their state representations are identical (this is a trivial conceptual extension to the Alice-Bob example from Section~\ref{sec:modelling}). $\theta^R$ and $\theta^T$ are not necessarily (and indeed are very likely not) uncorrelated, but we model them as distinct variables for clarity. For the same reasoning as was given with respect to $\theta^T$, influence links will also be necessary between $A_x$ and $\theta^R_{x+1}$.

Figure~\ref{fig3:tf_rf} depicts the media recommendation CID that results from these extensions to the MDP. To reinforce the point made in Subsection~\ref{sec:extracting_cid} about variations on the MDP not affecting the causal structure local to the variables $\{\theta^T_t | t \in \mathbb{N}\}$, we have highlighted these variables' incoming and outgoing causal links in the figure; the reader may compare these to those in Figure~\ref{fig2:tf_only} to verify that all the same links are present.

\section{Formal MDP Definition of the Recommendation Simulation}
\label{sec:app_c}

We define the MDP $\langle S,A,T,R,\gamma \rangle$ of the media recommendation problem described in Section~\ref{sec:experiment} as follows:

\begin{itemize}
    \item $S = \{(s^{LR}, s^{LC}, s^{CR}, s^{CC}, s^{RR}, s^{RC}) \in \mathbb{N}^6 \: | \: (s^{LR} + s^{CR} + s^{RR} \leq h) \land (s^{LR} \geq s^{LC}) \land (s^{CR} \geq s^{CC}) \land (s^{RR} \geq s^{RC})\}$.
    \begin{itemize}
        \item $s_t$ is the state after $t$ recommendations, $0 \leq t \leq h$.
        \item The state space is interpreted as follows:
        \begin{itemize}
            \item $s_t^{LR}$ is the number of "left-wing" recommendations made to the user after $t$ total recommendations
            \item $s_t^{LC}$ is the number of "left-wing" recommendations clicked on by the user after $t$ total recommendations
            \item $s_t^{CR}$ and $s_t^{CC}$ are as above, but with respect to "centrist" recommendations
            \item $s_t^{RR}$ and $s_t^{RC}$ are as above, but with respect to "right-wing" recommendations
        \end{itemize}
    \end{itemize}
    \item $A = \{0,1,2\}$, where:
    \begin{itemize}
        \item 0 = `Left-wing recommendation'
        \item 1 = `Centrist recommendation'
        \item 2 = `Right-wing recommendation'
    \end{itemize}
    \item $T$ is defined as follows, where $s = (s^{LR}, s^{LC}, s^{CR}, s^{CC}, s^{RR}, s^{RC})$:
    \begin{itemize}
        \item $T(s, 0, (s^{LR} + 1, s^{LC} + 1, s^{CR}, s^{CC}, s^{RR}, s^{RC}))$ = $\theta^{T^L}$
        \item $T(s, 0, (s^{LR} + 1, s^{LC}, s^{CR}, s^{CC}, s^{RR}, s^{RC}))$ = $(1 - \theta^{T^L})$
        \item $T(s, 1, (s^{LR}, s^{LC}, s^{CR} + 1, s^{CC} + 1, s^{RR}, s^{RC}))$ = $\theta^{T^C}$
        \item $T(s, 1, (s^{LR}, s^{LC}, s^{CR} + 1, s^{CC}, s^{RR}, s^{RC}))$ = $(1 - \theta^{T^C})$
        \item $T(s, 2, (s^{LR}, s^{LC}, s^{CR}, s^{CC}, s^{RR} + 1, s^{RC} + 1))$ = $\theta^{T^R}$
        \item $T(s, 2, (s^{LR}, s^{LC}, s^{CR}, s^{CC}, s^{RR} + 1, s^{RC}))$ = $(1 - \theta^{T^R})$
        \item $T(s,a,s) = 0$ otherwise.\footnote{Less formally, this transition function just amounts to the intuition that recommending a post from one source will increment the number of total recommendations from that source so far, and also increment the number of clicks on that source's posts with the relevant probability.}
        \end{itemize}
    \item $R(s_t, s_{t+1})$ is defined as: \[\begin{cases} 
      1 & (s_{t+1}^{LC} - s_t^{LC}) + (s_{t+1}^{CC} - s_t^{CC}) + (s_{t+1}^{RC} - s_t^{RC}) = 1 \\
      0 & \text{otherwise.}
   \end{cases}\]

   \item $\gamma = 0.999$
    
\end{itemize}

Note that this specific MDP interpretation of the media recommendation problem fits within our general MDP definition from Section~\ref{sec:modelling}.

Finally, we define the causal effect of agent actions on the user's exogenous variables in our simulation; this is not something that would be explicitly defined in a scenario with real users, but we need to define it here in order to build our simulation. As mentioned in Section~\ref{sec:experiment}, for this effect we took inspiration from recent research into user polarisation on social media, which has demonstrated that showing people who identify with one wing of the political spectrum volumes of content from the opposing wing can often increase user polarisation \citep{bail18, bail21}. We approximate this effect with the following causal relationship between the recommendation at time $t$, and the value of $\theta^T_{t+1}$:

\begin{itemize}
    \item If the user is right-wing, and $a_t$ = 0 (a left-wing recommendation), then $\theta^{T^R}_{t+1} = \text{min}(p\theta^{T^R}_t, 1.0)$ for some random variable $p \sim P$ where $\mathbb{E}[p]$ > 1.0 . We call $p$ the `polarization factor.'
    \item The same effect applies for left-wing users with $a_t$ = 2 and $\theta^{T^L}_{t+1}$.
\end{itemize}

\end{document}